\documentclass[10pt]{article} 
\usepackage[preprint]{tmlr}   

\usepackage{amsmath,amssymb,amsfonts}
\usepackage{graphicx}
\usepackage{booktabs}
\usepackage{multirow}
\usepackage{array}
\usepackage{siunitx}
\usepackage{microtype}
\usepackage{enumitem}
\usepackage{xcolor}
\usepackage{tikz}
\usetikzlibrary{arrows.meta,positioning}
\usepackage{float}
\usepackage[hidelinks]{hyperref}
\usepackage{url}

\newcommand{\Hu}{\ensuremath{H(u)}}
\newcommand{\Ha}{\ensuremath{H(a)}}
\newcommand{\ustar}{u^{\star}}

\newcommand{\pp}{\,\mathrm{pp}}
\newcommand{\Rstoch}{R_{\mathrm{stoch}}}
\newcommand{\Rdet}{R_{\mathrm{det}}}
\newcommand{\meanlogsig}{\overline{\log\sigma}}

\title{Where Entropy Is Measured Matters: Policy Geometry in Bounded Continuous-Control PPO}

\author{\name Yiyang He\thanks{Equal contribution.} \email y.he20@lancaster.ac.uk\\
\addr Department of Engineering, Lancaster University, Lancaster, UK
\AND
\name Zhichun Zhou\footnotemark[1] \email zy2332222@buaa.edu.cn\\
\addr Department of Engineering, Beihang University, Beijing, China
\AND
\name Ziwei Wang \email z.wang82@lancaster.ac.uk\\
\addr  Department of Engineering, Lancaster University, Lancaster, UK
\AND
\name Tao Xue\thanks{Corresponding author: xuetao16@mail.tsinghua.edu.cn.} \email xuetao16@mail.tsinghua.edu.cn\\
\addr Department of Automation, Tsinghua University, Beijing, China
\AND
\name Haolin Fei \email haolin.fei@ieee.org\\
\addr Department of Engineering, Lancaster University, Lancaster, UK}
\def\month{MM}  
\def\year{YYYY} 

\begin{document}
\maketitle

\begin{abstract}
Many continuous-control policies are optimized as unbounded Gaussians and then mapped into bounded actions. We show that the space in which entropy is measured changes the geometry that proximal policy optimization (PPO) learns. The study begins with an 80-muscle MyoLeg walking task, where a clipped Gaussian executes 89.07\% of actions within 5\% of a bound. A counterfactual decomposition on the same visited states shows that this is not an effect of variance alone. Setting the variance to zero still leaves 83.83\% of actions near a bound, centering the mean while keeping the learned variance gives 77.51\%, and 82.12\% of state-conditioned means lie outside the executable interval. Replacing clipping by a smooth tanh map does not remove the high-variance regime. For a latent Gaussian, the latent entropy \Hu{} has zero gradient with respect to the mean and a constant variance-increasing gradient of the entropy loss. For the executed-action entropy \Ha{}, the transformation Jacobian adds $2c_{\mathrm{ent}}\mathbb{E}[\tanh u_i]$ to the entropy-loss gradient on the mean; gradient descent therefore pulls the mean inward. Across three matched MyoLeg seeds, total near-boundary occupancy is 71.42\%, 29.76\%, and 18.83\% under latent entropy, no entropy, and executed-action entropy. No entropy gives the lowest occupancy when only variance is retained, at the 5\% reference margin, yet \Ha{} produces the most interior policy mean, so low learned dispersion is not equivalent to interior policy geometry. A 38-dimensional Dog-Stand replication with an independent, CleanRL-based PPO implementation reproduces the ordering in mean geometry, which also survives evaluation on shared states and boundary margins from 1\% to 10\%. Direct mean penalties reproduce or exceed the centering produced by \Ha{}, so interior means are not unique to executed entropy. However, an L2-regularized policy and \Ha{} reach nearly identical mean geometry while differing by 0.600 in mean log standard deviation and by 129 return points in Dog-Stand. Entropy measurement space is therefore a coupled mean--variance design choice, and task return alone does not characterize the geometry of a bounded policy.
\end{abstract}

\section{Introduction}
Continuous-control reinforcement learning often uses a Gaussian policy even when every actuator is bounded. The policy therefore lives in two spaces. In the \emph{latent} space, an action is sampled from an unbounded distribution. In the \emph{executed} space, that sample is clipped or transformed before it reaches the simulator or robot. This distinction is easy to ignore when return is the only evaluation metric. It becomes important when the learned controller spends much of its time near the action limits.

We encountered this problem in an 80-muscle MyoLeg walking task. An off-the-shelf PPO configuration produced high-return locomotion, yet the muscle policy was strongly boundary dominated: roughly 89\% of executed muscle commands were within 5\% of either 0 or 1. A 14-actuator motor policy on the same model showed only about 9.5\% near-boundary occupancy, measured with the same absolute 0.05 margin from either bound (Section~\ref{sec:geometry}). The immediate explanations were physical: perhaps muscle activation filtering, asymmetric activation/deactivation time constants, or the non-negative muscle action range caused the switching. They did not. Removing the activation filter or making the time constants symmetric left the regime essentially unchanged.

The more important question is statistical: \emph{is the policy near the bounds because its variance is large, or because its state-conditioned mean has moved there?} We answer this with a same-state decomposition. In the trained clipped-Gaussian muscle policy, setting the variance to zero still leaves 83.8\% near-boundary occupancy, and 82.1\% of state-conditioned means lie outside the executable interval. Large dispersion is also sufficient, but it is not the whole explanation. The policy mean itself is strongly extruded. We use \emph{mean extrusion} to denote this movement of a state-conditioned latent mean into regions that the bounded execution map sends close to an action limit.

This observation points to the entropy term. For a diagonal latent Gaussian $u\sim\mathcal{N}(\mu,\sigma^2)$,
\begin{equation}
H(u)=\sum_i\left(\log\sigma_i+\tfrac{1}{2}\log(2\pi e)\right),
\end{equation}
so latent entropy has no direct gradient on the mean and gives the same variance-increasing entropy-loss gradient in every dimension. If entropy is measured after a bounded transform, the change-of-variables Jacobian adds a geometric term. For the tanh map, this term pulls the mean inward and changes the variance gradient from a constant into a distribution-dependent quantity.

We test this mechanism with three matched tanh policies: latent entropy \Hu{}, no entropy, and executed-action entropy \Ha{}. The intervention changes only the entropy objective; sampling and the PPO importance ratio remain in the same latent Gaussian family. Across three MyoLeg seeds, the three policies converge to sharply different mean--variance regimes. Latent entropy has the largest dispersion and the most extruded means. Removing entropy drives dispersion much lower but leaves the mean substantially closer to the bounds than \Ha{}. Thus the policy with the smallest exploration-driven boundary occupancy does not have the smallest total boundary occupancy.

We then repeat the experiment on the 38-dimensional DeepMind Control Dog-Stand task using an independent, CleanRL-based PPO implementation. The same ordering in mean geometry appears across all three seeds and three late checkpoints. It also survives two robustness checks that target natural alternative explanations: every policy is scored on shared state buffers, and the near-boundary margin is swept from 1\% to 10\%. The same Dog study also changes the interpretation of \Ha{}. Two direct mean penalties produce equally or more interior means. In particular, a pre-tanh L2 penalty nearly matches the trained mean distribution of \Ha{}, yet the two policies end in very different variance and return regimes. Executed-action entropy is therefore not the only route to an interior mean; it is a coupled mean--variance regularizer.

The contribution is mechanistic rather than a new entropy formula. Tanh-transformed densities and Jacobian-corrected log probabilities are standard, including in Soft Actor-Critic (SAC)~\citep{haarnoja2018sac}. Our contribution is to connect the choice of entropy measurement space to trained PPO policy geometry, separate mean placement from dispersion experimentally, and test the resulting mechanism with direct gradients and an independent external replication.

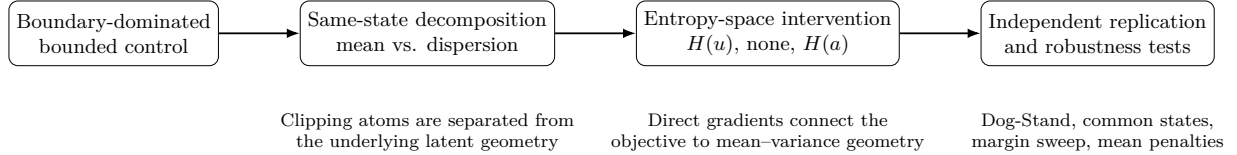
\begin{figure}[H]
\centering
\resizebox{0.98\textwidth}{!}{%
\begin{tikzpicture}[
node distance=1.4cm and 1.25cm,
box/.style={draw,rounded corners,align=center,minimum width=3.0cm,minimum height=1.0cm,font=\small},
arr/.style={-{Latex[length=2mm]},thick}
]
\node[box] (obs) {Boundary-dominated\\bounded control};
\node[box,right=of obs] (dec) {Same-state decomposition\\mean vs. dispersion};
\node[box,right=of dec] (ent) {Entropy-space intervention\\\Hu{}, none, \Ha{}};
\node[box,right=of ent] (gen) {Independent replication\\and robustness tests};
\draw[arr] (obs) -- (dec);
\draw[arr] (dec) -- (ent);
\draw[arr] (ent) -- (gen);
\node[align=center,font=\footnotesize,below=0.6cm of dec] {Clipping atoms are separated from\\the underlying latent geometry};
\node[align=center,font=\footnotesize,below=0.6cm of ent] {Direct gradients connect the\\objective to mean--variance geometry};
\node[align=center,font=\footnotesize,below=0.6cm of gen] {Dog-Stand, common states,\\margin sweep, mean penalties};
\end{tikzpicture}}
\caption{Study logic. The paper starts from an actuator-level phenomenon, separates mean placement from dispersion, changes only the space in which entropy is measured, measures the resulting gradients, and then tests the mechanism in an independent PPO implementation.}
\label{fig:overview}
\end{figure}

The main contributions are:
\begin{itemize}[leftmargin=1.25em]
\item A counterfactual decomposition evaluated on the same states, showing that mean placement and dispersion are each sufficient to produce high boundary occupancy in trained 80-muscle PPO, and that the policy mean itself is strongly extruded.
\item Controlled ablations showing that muscle activation filtering, activation/deactivation asymmetry, and hard clipping are not necessary explanations for the boundary-dominated regime.
\item A three-condition intervention on entropy measurement space, replicated across three MyoLeg seeds, together with analytic gradients and measurements on frozen buffers that directly distinguish latent from executed entropy.
\item An independent 38-dimensional Dog-Stand replication under CleanRL PPO, followed by an audit on shared states and a sweep of boundary margins from 1\% to 10\%.
\item Controls with direct penalties on the policy mean, showing that interior means are not unique to \Ha{}, while closely matched mean geometry can coexist with very different variance and performance regimes.
\end{itemize}
Figure~\ref{fig:overview} summarizes this chain of reasoning.

\section{Related Work}
\subsection{Bounded continuous-control policies}
The mismatch between an unbounded Gaussian and a bounded action space is well known. Beta policies place probability directly on a bounded interval~\citep{chou2017beta}, including prior PPO-specific evaluations of Beta policies on bounded action spaces~\citep{petrazzini2021beta}. Clipped-action policy gradient corrects the policy-gradient treatment of clipping~\citep{fujita2018clipped}, marginal policy gradients unify a family of estimators for clipped and directional bounded-action policies~\citep{eisenach2019marginal}, and tanh-squashed Gaussians are standard in SAC~\citep{haarnoja2018sac}. These methods establish that action parameterization matters. Our question is different: when PPO uses a latent Gaussian and a bounded execution map, how do the entropy term, mean placement, and variance jointly shape the trained controller?

SAC is particularly relevant because its squashed-Gaussian actor evaluates the transformed action density, including the tanh Jacobian~\citep{haarnoja2018sac}. In the terminology of this paper, its entropy term is evaluated in the executed action space. We do not claim the change-of-variables identity or \Ha{} as new. We use the identity to expose a geometric consequence of a common PPO convention: evaluating the entropy of the latent Gaussian before the action is bounded. The convention is common but not universal: the Brax training stack estimates its PPO entropy bonus on the tanh-squashed action, including the change-of-variables correction~\citep{freeman2021brax}.

Boundary-preferring policies can also perform well. \citet{seyde2021bangbang} showed that bang-bang and Bernoulli policies solve a range of continuous-control tasks. That result makes return an incomplete description of a controller. We focus on how a Gaussian PPO policy acquires boundary-heavy geometry rather than on whether boundary control can be optimal for a task. From a similar starting point, \citet{lin2025gac} argues that squashing a Gaussian into a bounded interval distorts the geometry of the action space and creates gradient pathologies near the saturated boundaries, and responds by replacing the distribution itself with a spherical direction--concentration parameterization. We keep the squashed-Gaussian parameterization fixed instead and intervene only on the space in which the PPO entropy bonus is measured; the same-state decomposition then attributes the boundary-heavy executed distribution to mean placement and dispersion separately.

\subsection{Entropy, variance, and structured exploration}
PPO~\citep{schulman2017ppo} often uses an entropy bonus to maintain stochasticity, a device that dates to REINFORCE-style policy search~\citep{williams1991entropy} and became standard practice through asynchronous actor--critic methods~\citep{mnih2016a3c}. A separate line of work treats entropy not as a bonus but as part of the objective itself, in the maximum-entropy formulation~\citep{ziebart2008maxent,haarnoja2017softq} of which SAC is the continuous-control instance~\citep{haarnoja2018sac}. Large empirical studies show that implementation choices such as state-independent versus state-dependent standard deviations can materially affect on-policy learning~\citep{andrychowicz2021what}. Code-level details of PPO can likewise dominate its reported advantage over alternatives~\citep{engstrom2020implementation}. Our result is of that kind: the choice of coordinate system for the entropy term is an implementation convention with a specific, measurable geometric consequence. \citet{zhou2026gaussian} study Gaussian variance in PPO and propose an explicit variance-management procedure. Other work changes the \emph{structure} of exploration rather than only its marginal variance. Generalized state-dependent exploration (gSDE) introduces state-dependent, temporally persistent exploration for robotic control~\citep{raffin2022gsde}. Pink-noise exploration introduces temporal correlation in continuous-action noise~\citep{eberhard2023pink}, and Lattice adds structured correlation in time and actuator space for overactuated systems~\citep{chiappa2023lattice}. Our experiments hold these choices fixed and isolate a different variable: whether the entropy objective is evaluated before or after the bounded action transform.

\subsection{Action regularization and smooth control}
Action regularization provides another route to changing controller geometry. CAPS explicitly regularizes the policy to produce smoother state-to-action mappings and control sequences~\citep{mysore2021caps}. Our tanh-squared and pre-tanh L2 penalties are not proposed as new smooth-control methods. They are deliberately simple, per-state mean regularizers used as mechanism controls: they test whether the mean-centering effect of executed entropy can be reproduced without changing the entropy definition. Unlike CAPS, they do not penalize temporal action differences or local state-to-action sensitivity.

\subsection{Musculoskeletal control}
MyoSuite and related neuromusculoskeletal benchmarks expose many bounded muscle activations on one articulated body~\citep{caggiano2022myosuite,kidzinski2018learning,caggiano2024myochallenge}. Prior work addresses exploration and control in these high-dimensional systems with specialized exploration and control architectures~\citep{schumacher2022deprl,chiappa2023lattice,lee2019scalable}. Muscle activation dynamics are also known to affect optimized trajectories~\citep{anderson2001dynamic,vandenbogert2025}. We therefore test the muscle dynamics directly before attributing the observed boundary behavior to the RL objective.

\section{Policy Geometry in a Bounded Action Space}
\label{sec:geometry}
\subsection{Latent and executed actions}
Let PPO produce a diagonal Gaussian latent action
\begin{equation}
u=\mu_\theta(s)+\sigma\odot\epsilon,\qquad \epsilon\sim\mathcal{N}(0,I),
\label{eq:latent}
\end{equation}
with state-independent $\log\sigma$ in the implementations studied here. The environment executes $a=f(u)$. We use either hard clipping or a tanh map. For the muscle action interval $[0,1]$,
\begin{equation}
a=\frac{\tanh u+1}{2}.
\end{equation}
A 5\% near-boundary margin, $a\le0.05$ or $a\ge0.95$, is equivalent to
\begin{equation}
\begin{aligned}
|\tanh u|&\ge0.9 \Longleftrightarrow |u|\ge\ustar,\\
\ustar&=\operatorname{atanh}(0.9)=1.4722194896.
\end{aligned}
\label{eq:ustar}
\end{equation}
Dog-Stand uses the normalized coordinate $y=\tanh u\in(-1,1)$ and the same criterion $|y|\ge0.9$.

Two margin conventions appear in this paper and are not interchangeable on a symmetric action box. The clipped-Gaussian battery of Section~\ref{sec:isolation} measures an \emph{absolute} distance of 0.05 from either bound of the configuration's own box, which is 5\% of the muscle range $[0,1]$ and 2.5\% of the motor range $[-1,1]$. Every tanh and Dog-Stand quantity instead uses a \emph{fraction of the action range}, giving $|y|\ge1-2m$ with $m=0.05$ by default. The two coincide on $[0,1]$, so all muscle numbers are directly comparable. The motor reference in Table~\ref{tab:isolation} is the one row measured at a stricter relative threshold; it is reported this way because it is the interior reference rather than part of the mechanism comparison.

\subsection{Same-state mean--variance decomposition}
Near-boundary occupancy can come from the mean, the variance, or both. We evaluate four counterfactual cells on the same visited states:
\begin{align}
P_{11}&:\ (\mu,\sigma)\text{ actual},\\
P_{10}&:\ (\mu,0)\text{ mean only},\\
P_{01}&:\ (\mu_c,\sigma)\text{ variance only},\\
P_{00}&:\ (\mu_c,0)\text{ centered baseline},
\end{align}
where $\mu_c=0.5$ for clipping to $[0,1]$ and $\mu_c=0$ for the symmetric tanh latent coordinate. For tanh,
\begin{equation}
P_{11}=\Phi\!\left(\frac{-\ustar-\mu}{\sigma}\right)
+1-\Phi\!\left(\frac{\ustar-\mu}{\sigma}\right),
\label{eq:p11}
\end{equation}
computed componentwise and averaged over states and action dimensions. The mean-only cell is $\mathbf{1}(|\mu|\ge\ustar)$. The variance-only cell uses the same formula with $\mu=0$.

For clipping to $[0,1]$, the near-boundary thresholds are 0.05 and 0.95. We separately report the \emph{exact atom mass}, which uses thresholds 0 and 1. The interaction
\begin{equation}
I=P_{11}-P_{10}-P_{01}+P_{00}
\end{equation}
is large and negative when mean and variance are redundant under the 100\% ceiling. We therefore report the four cells directly rather than treating an additive attribution as unique.

\subsection{Latent and executed entropy}
For the latent Gaussian,
\begin{equation}
\frac{\partial H(u)}{\partial\mu_i}=0,
\qquad
\frac{\partial H(u)}{\partial\log\sigma_i}=1.
\label{eq:latentgrad}
\end{equation}
With entropy loss $L_{\mathrm{ent}}=-c_{\mathrm{ent}}H$, latent entropy contributes
\begin{equation}
\frac{\partial L_{\mathrm{ent}}}{\partial\log\sigma_i}=-c_{\mathrm{ent}}
\end{equation}
for every action dimension, independent of the current variance. It contributes no direct mean force.

For an invertible differentiable transform $a=f(u)$,
\begin{equation}
H(a)=H(u)+\mathbb{E}_{u}\!\left[\log|\det J_f(u)|\right].
\label{eq:ha}
\end{equation}
For the componentwise map $a_i=(\tanh u_i+1)/2$, reparameterization gives
\begin{align}
\frac{\partial H(a)}{\partial\mu_i}
&=-2\,\mathbb{E}[\tanh u_i],
\label{eq:hamean}\\
\frac{\partial H(a)}{\partial\log\sigma_i}
&=1-2\sigma_i^2\mathbb{E}[\operatorname{sech}^2u_i].
\label{eq:havar}
\end{align}
Thus the gradient of the entropy loss with respect to the mean is $+2c_{\mathrm{ent}}\mathbb{E}[\tanh u_i]$. Gradient descent moves positive means down and negative means up: the direct effect is inward. The variance term is finite and distribution dependent rather than constant.

The bounded transform does not need to appear in the PPO importance ratio. At a stored latent sample $u$,
\begin{equation}
\frac{\pi_{\mathrm{new}}(a)}{\pi_{\mathrm{old}}(a)}
=
\frac{p_{\mathrm{new}}(u)/|\det J_f(u)|}
{p_{\mathrm{old}}(u)/|\det J_f(u)|}
=
\frac{p_{\mathrm{new}}(u)}{p_{\mathrm{old}}(u)}.
\label{eq:ratio}
\end{equation}
For the ideal invertible tanh transform, the Jacobian cancels exactly at the stored latent samples. Our \Ha{} intervention therefore changes the entropy calculation but not the latent Gaussian sampling rule or PPO ratio; the MyoLeg implementation adds an inert $\pm10$ latent interface clip, quantified in Section~\ref{sec:methods}. Any fixed affine rescaling applied after the bounded map changes \Ha{} only by an additive constant that is independent of the policy parameters, so it leaves every entropy gradient unchanged.

\section{Experimental Design}
\label{sec:methods}
\subsection{MyoLeg task and PPO training}
The primary task is a hybrid MyoLeg model based on MyoSuite~\citep{caggiano2022myosuite} and MuJoCo~\citep{todorov2012mujoco}. The model contains 80 muscle actuators and 14 motor actuators on the same skeleton. Muscle commands lie in $[0,1]^{80}$; motor commands lie in $[-1,1]^{14}$. The observation has 101 dimensions. The physics timestep is 1\,ms and the environment applies a frame skip of 4, so the control frequency is 250\,Hz; both numbers were read back from the live environment and are recorded in the release manifest. Episodes are limited to 1000 control steps and can terminate early on a pelvis-height or orientation violation.

The reference motion is CMU clip \texttt{02\_01.c3d}~\citep{cmumocap}, with a 133-frame gait cycle at 120 Hz and a walking speed of approximately 1.187 m/s. The implemented reward is
\begin{equation}
\begin{aligned}
r &= 20\,r_{\mathrm{ee}}\left(0.75r_q+0.10r_v\right),\\
r_q &= \exp\!\left[-0.857\sum_{j=1}^{14}(q_j^{\star}-q_j)^2\right],\\
r_{\mathrm{ee}} &= \exp\!\left[-40\sum_k(e_k^{\star}-e_k)^2\right],\qquad r_v\equiv 1.
\end{aligned}
\label{eq:myo_reward}
\end{equation}
The velocity subreward is constant because the environment implementation sets the velocity-difference variable used by that term to zero. Thus it contributes no velocity-tracking information; it contributes $20\times0.10\,r_{\mathrm{ee}}=2r_{\mathrm{ee}}$ to the scalar reward, and the per-step reward ceiling is $20(0.75+0.10)=17$ when the tracking terms equal one. This is an environment-definition quirk rather than an experimental condition. It can affect the task being learned, but the same reward code is used in every MyoLeg entropy condition and therefore cannot explain their matched between-condition differences.

The MyoLeg PPO experiments use Stable-Baselines3~\citep{raffin2021sb3}. The primary tanh comparison uses separate 512--512 ReLU policy and value networks, learning rate $10^{-4}$, rollout length 2048 per environment, minibatch size 128, 10 epochs, $\gamma=0.99$, generalized advantage estimation~\citep{schulman2016gae} with $\lambda=0.99$, clip range 0.2, and 12 parallel environments. Both implementations optimize with Adam~\citep{kingma2015adam}. The latent action interface is widened to $[-10,10]^{80}$; clipping can still occur in the saturated tails under \Hu{}, but its effect on the executed action is numerically negligible, as quantified next. Measured on canonical late-training rollouts, a Gaussian sample exceeds the box in $3.9\%\pm0.7\%$ of components, and at least one of the 80 components exceeds it on $95.9\%\pm2.5\%$ of control steps ($n=3$ seeds; analytic tail probabilities and empirical counts agree to three decimals). With no entropy the component-level rate is at most $1.1\times10^{-9}$ in any seed, and under \Ha{} it is zero at double precision; in particular, the one condition whose entropy objective contains the transformation Jacobian never evaluates it at a clipped sample. Where the clip does occur it is inert for two reasons. First, Stable-Baselines3 clips only the copy of the action passed to the environment; the rollout buffer stores the unclipped latent sample, so the importance ratio and every entropy quantity are computed exactly as analyzed. Second, $\tanh 10 = 1-4.1\times10^{-9}$, so a clipped component's executed activation differs from the unclipped ideal by at most $2.1\times10^{-9}$ on the $[0,1]$ scale, four orders of magnitude below every threshold used in this paper. All three primary conditions are trained for 145,022,976 environment steps with training seeds 0, 1, and 2. Appendix~\ref{app:config} gives the full training and network configuration for both implementations.

The three entropy conditions are:
\begin{enumerate}[leftmargin=1.45em]
\item \textbf{Latent entropy \Hu}: $c_{\mathrm{ent}}=10^{-3}$ and the ordinary Gaussian entropy.
\item \textbf{No entropy}: $c_{\mathrm{ent}}=0$.
\item \textbf{Executed-action entropy \Ha}: $c_{\mathrm{ent}}=10^{-3}$ and Eq.~\eqref{eq:ha}, estimated with four antithetic reparameterized samples per state.
\end{enumerate}
The latent Gaussian family and PPO ratio are otherwise identical.

Before the tanh intervention, five configurations isolate physical and distributional explanations: a 14-actuator motor policy; an 80-muscle activation-override policy; the native muscle activation filter; a symmetric activation-time-constant variant; and a bounded Beta policy~\citep{chou2017beta}. These experiments use three training seeds each and matched canonical evaluation. Their role is diagnostic: they establish which ingredients are not necessary for the original clipped-Gaussian regime.

\subsection{Dog-Stand external replication}
The external task is \texttt{dm\_control/dog-stand-v0}~\citep{tassa2020dmcontrol}, accessed through the Gymnasium API~\citep{towers2024gymnasium} via the Shimmy compatibility layer~\citep{shimmy2023}, with a 223-dimensional observation and a 38-dimensional bounded action. We start from CleanRL's single-file continuous-action PPO implementation~\citep{huang2022cleanrl}. For the three primary entropy conditions, the action transform, entropy objective, and diagnostic logging are the only algorithmic changes. The clipped surrogate, generalized advantage estimation~\citep{schulman2016gae}, value loss, advantage normalization, minibatching, Adam optimizer~\citep{kingma2015adam}, and the remaining hyperparameters retain the CleanRL configuration listed in Appendix~\ref{app:config}. The controls with direct mean penalties, introduced after the primary experiment, add the stated actor-loss penalty and otherwise use the same configuration.

The policy samples $u\sim\mathcal{N}(\mu,\sigma)$, executes $y=\tanh u$, and then applies the environment's affine action scaling. The three primary entropy conditions use the same $c_{\mathrm{ent}}$ values as MyoLeg without tuning. Each condition is trained from scratch for 3,000,320 steps with seeds 0, 1, and 2. Dog-Stand episodes last 15 s, or 1000 control steps at 0.015 s per step (66.67 Hz).

The canonical evaluator runs 30 stochastic and 30 deterministic episodes. Reset seeds and action-noise seeds are paired across policies; the checkpoint's observation normalizer is restored and frozen; returns are raw environment returns. Same-state $P_{11}$, $P_{10}$, and $P_{01}$ are evaluated on states visited by the stochastic policy. Deterministic-rollout occupancy is reported separately because deterministic execution visits a different state distribution.

Three geometry criteria were specified before the final Dog runs. First, the late-training \Hu{} variance should remain above no entropy and \Ha{}. Second, \Ha{} should reduce mean extrusion relative to both alternatives. Third, no entropy should have lower $P_{01}$ than \Ha{} while retaining higher total occupancy. Return was not a criterion.

\subsection{Direct mean regularization controls}
After the primary Dog experiment, we added two controls to ask whether \Ha{} is simply a complicated way to center the mean:
\begin{align}
L_{\tanh^2}&=\frac{\lambda_{\tanh^2}}{2}\,\mathbb{E}_s\!\left[\sum_i\tanh^2\mu_i(s)\right],\\
L_{\mu^2}&=\frac{\lambda_{\mu^2}}{2}\,\mathbb{E}_s\!\left[\sum_i\mu_i(s)^2\right].
\end{align}
The batch operation is a mean over states and a sum over action dimensions. Both are trained with no entropy. Their coefficients are fixed by matching the magnitude of the \Ha{} entropy-loss gradient with respect to the mean, taken in the zero-dispersion limit at the 5\% boundary threshold:
\begin{equation}
2c_{\mathrm{ent}}\tanh\ustar=1.8\times10^{-3},
\end{equation}
which gives $\lambda_{\tanh^2}=0.0105263158$ and $\lambda_{\mu^2}=0.0012226438$. Each control is trained for 3,000,320 steps with the same three Dog seeds and evaluated with the same canonical protocol.

\subsection{Gradient diagnostics}
For MyoLeg, frozen-gradient decomposition is performed on the late-training seed-0 checkpoint of each entropy condition using 10 independently collected on-policy buffers per checkpoint. We decompose the actor loss into the PPO surrogate and entropy terms, separately for $\mu$ and $\log\sigma$. Buffer fingerprints are checked for uniqueness before cross-buffer statistics are computed. The entropy-side mean force is summarized by $\langle g_{\mu,\mathrm{ent}},\mu\rangle$ and, when the gradient is nonzero, by its cosine with $\mu$. Under \Hu{} and no entropy the entropy mean-gradient vector is exactly zero, so the cosine is undefined rather than zero.

For Dog, the inward mean mechanism is evaluated on 30 independent buffers per late-training checkpoint and seed. The terminal log-variance decomposition is also measured, but its confidence intervals span zero under all three primary conditions; we therefore use it only to bound the interpretation of the long-horizon variance trajectories.

\subsection{Robustness analyses}
The \textbf{common-state audit} tests whether the ordering in mean geometry is caused by the policies visiting different states. For each Dog seed, all five policies donate 10 episodes of raw pre-normalization states. Every trained policy is scored on every donor set and on an equal-mixture pool. Each evaluated policy applies its own frozen observation normalizer before the forward pass.

The \textbf{boundary-margin analysis} recomputes $P_{11}$, $P_{10}$, and $P_{01}$ analytically at margins of 1\%, 2.5\%, 5\%, and 10\% on one freshly collected 30-episode on-policy state set per checkpoint. All four thresholds use the same states within a sweep. Both tasks use the same NumPy action-noise convention as their frozen canonical evaluators, and both reproduce the corresponding 5\% cells of Tables~\ref{tab:myo3} and \ref{tab:dog5} exactly. Appendix~\ref{app:margin} reports the sweep values.

\subsection{Statistical units}
The training seed is the unit for method-level comparisons. Tables report the mean and sample standard deviation over three trained policies. Evaluation episodes characterize a fixed policy and are paired by reset and action-noise seeds where applicable; they are not treated as independent training replicates. Frozen gradient buffers are the unit only for local gradient estimation. With three training seeds we emphasize effect sizes and direction agreement rather than small-sample $p$-values. In particular, Dog return comparisons are descriptive: we report the three-seed mean, sample SD, and matched-seed direction, but make no population-level superiority claim from $n=3$.

\subsection{Code and evaluation artifacts}
The training and analysis code should be treated as part of the experimental specification. An archival code package accompanies this paper, containing the MyoLeg/SB3 and Dog/CleanRL training implementations, the \Ha{} estimator and its numerical checks, and the canonical evaluators used for the reported tables. The package also includes the audit on shared states, the sweep over boundary margins, gradient diagnostics, frozen result JSONs, and manifests with checkpoint and script SHA-256 hashes. The same tagged release, together with the Dog checkpoints shipped in the package and the hashed MyoLeg checkpoints, will be maintained in a permanent public repository. The release will also include an environment lockfile and system-version manifest; exact package versions are not inferred from checkpoint files. Third-party model and motion assets will be redistributed only where their licenses permit, otherwise the release will provide acquisition and checksum instructions. The release manifest identifies the canonical evaluator that is the source of record for each table rather than asking readers to reproduce results with ad-hoc rollout scripts.

\section{Results}
\label{sec:results}
\subsection{The original boundary regime is not a muscle-dynamics effect}
\label{sec:isolation}
Table~\ref{tab:isolation} summarizes the initial mechanism-isolation experiments. The motor policy remains mostly interior, whereas all three clipped-Gaussian muscle variants spend roughly 87--89\% of their actions near a bound. Bypassing activation dynamics does not remove the regime, and making activation and deactivation time constants symmetric does not remove it either. The bounded Beta policy has much lower boundary occupancy and competitive stochastic return, showing that exact clipping atoms are not required for successful control on the task.

\subsection{Mean extrusion and dispersion are both sufficient}
\begin{figure}[H]
\centering
\includegraphics[width=\textwidth]{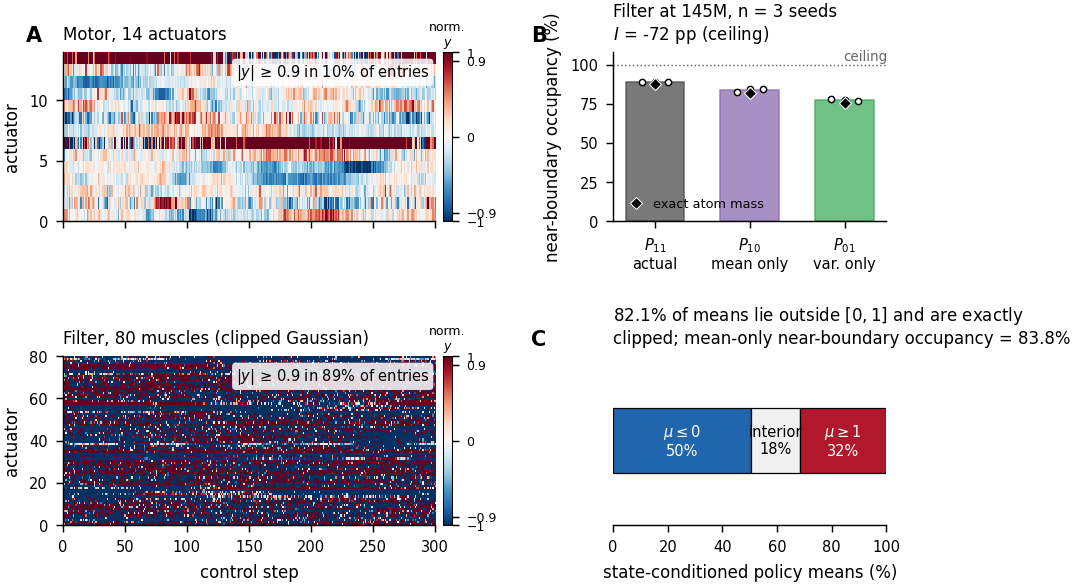}
\caption{The original MyoLeg phenomenon and its decomposition. \textbf{A:} Fixed-reset-seed traces for the first 300 control steps; no segment was selected after inspection. Motor actions are already in $[-1,1]$. Muscle actions $a\in[0,1]$ are mapped to $y=2a-1$ only for display, so $|y|\ge0.9$ is the same 5\% boundary margin. The 10\% and 89\% annotations describe these single traces, not the multi-seed canonical aggregates. The Motor annotation is therefore at the like-for-like 5\%-of-range threshold ($|a|\ge0.90$) noted in the Table~\ref{tab:isolation} caption, not at the absolute 0.05 margin used in that table; the canonical aggregate at this matching threshold is $10.12\%\pm0.06\%$, so the closeness of the single-trace value to the table's stricter-threshold 9.49\% is coincidental. \textbf{B:} Same-state Filter decomposition at 145M steps, $n=3$ training seeds. Bars show near-boundary occupancy: $P_{11}=89.07\%$, $P_{10}=83.83\%$, and $P_{01}=77.51\%$. Black diamonds show exact clipping-atom mass: 87.89\%, 82.12\%, and 75.61\%. The large negative interaction reflects overlap under the 100\% ceiling. \textbf{C:} 82.1\% of state-conditioned Filter means lie outside $[0,1]$ and are exactly clipped; the mean-only cell $P_{10}$ is 83.8\%.}
\label{fig:phenomenon}
\end{figure}

Figure~\ref{fig:phenomenon}B separates the two sources of boundary occupancy on the same Filter states. The actual policy gives $P_{11}=89.065\%\pm0.102\%$. Setting the variance to zero still gives $P_{10}=83.825\%\pm1.103\%$. Centering the mean while retaining the learned variance gives $P_{01}=77.515\%\pm0.428\%$. The interaction is about $-72\pp$, so the two causes are strongly redundant near the ceiling.

The mean is not merely close to the interval endpoints. Across the three Filter seeds, $82.12\%\pm1.26\%$ of state-conditioned means lie outside $[0,1]$. Consequently, reducing exploration variance alone cannot make this controller interior. Hard clipping adds exact atoms, but the latent policy geometry is already boundary oriented before clipping.

\begin{table}[H]
\centering
\caption{Initial MyoLeg mechanism-isolation configurations. Values are mean $\pm$ sample SD over three training seeds. Near-boundary occupancy uses an \emph{absolute} margin of 0.05 from either bound of the configuration's own action box, which is 5\% of the muscle range $[0,1]$ but 2.5\% of the motor range $[-1,1]$; the motor row is therefore measured at a stricter relative threshold than the muscle rows, and a like-for-like 5\%-of-range motor threshold is $|a|\ge0.90$; the same canonical protocol gives $10.12\%\pm0.06\%$ at that threshold. All five configurations use the same entropy coefficient $c_{\mathrm{ent}}=10^{-3}$: the four Gaussian configurations on their latent Gaussian entropy $H(u)$, and the Beta control on the entropy of its Beta distribution. The motor reference uses its prescribed 78{,}000{,}000-step budget. For the Beta policy, log disp.\ is an effective log-dispersion: the state-averaged log of the per-dimension Beta standard deviation implied by the learned concentrations. Beta uses an effective log-dispersion and is not numerically identical to Gaussian $\log\sigma$.}
\label{tab:isolation}
\scriptsize
\setlength{\tabcolsep}{2.8pt}
\begin{tabular}{lrrr}
\toprule
Condition & $\Rstoch$ & Near bound (\%) & log disp. \\
\midrule
Motor, 14D & $12637\pm941$ & $9.49\pm0.10$ & $-1.594\pm0.030$ \\
Override, 80D & $13437\pm730$ & $87.38\pm0.29$ & $+0.832\pm0.037$ \\
Filter, 80D & $12389\pm435$ & $89.07\pm0.16$ & $+0.798\pm0.029$ \\
Symmetric $\tau$, 80D & $12072\pm250$ & $88.73\pm0.45$ & $+0.852\pm0.018$ \\
Beta, 80D & $13698\pm294$ & $15.28\pm0.30$ & $-1.938\pm0.008$ \\
\bottomrule
\end{tabular}
\end{table}

\subsection{Entropy measurement space separates the trained tanh policies}
Hard clipping is not required for the high-variance latent regime. After replacing clipping by a smooth tanh action map, all three \Hu{} runs end in a much higher variance regime than their matched no-entropy and \Ha{} counterparts. We do not use a clip-versus-tanh slope ratio because the available cross-parameterization slope comparison is not resolved beyond seed variation.

\begin{figure}[H]
\centering
\includegraphics[width=\textwidth]{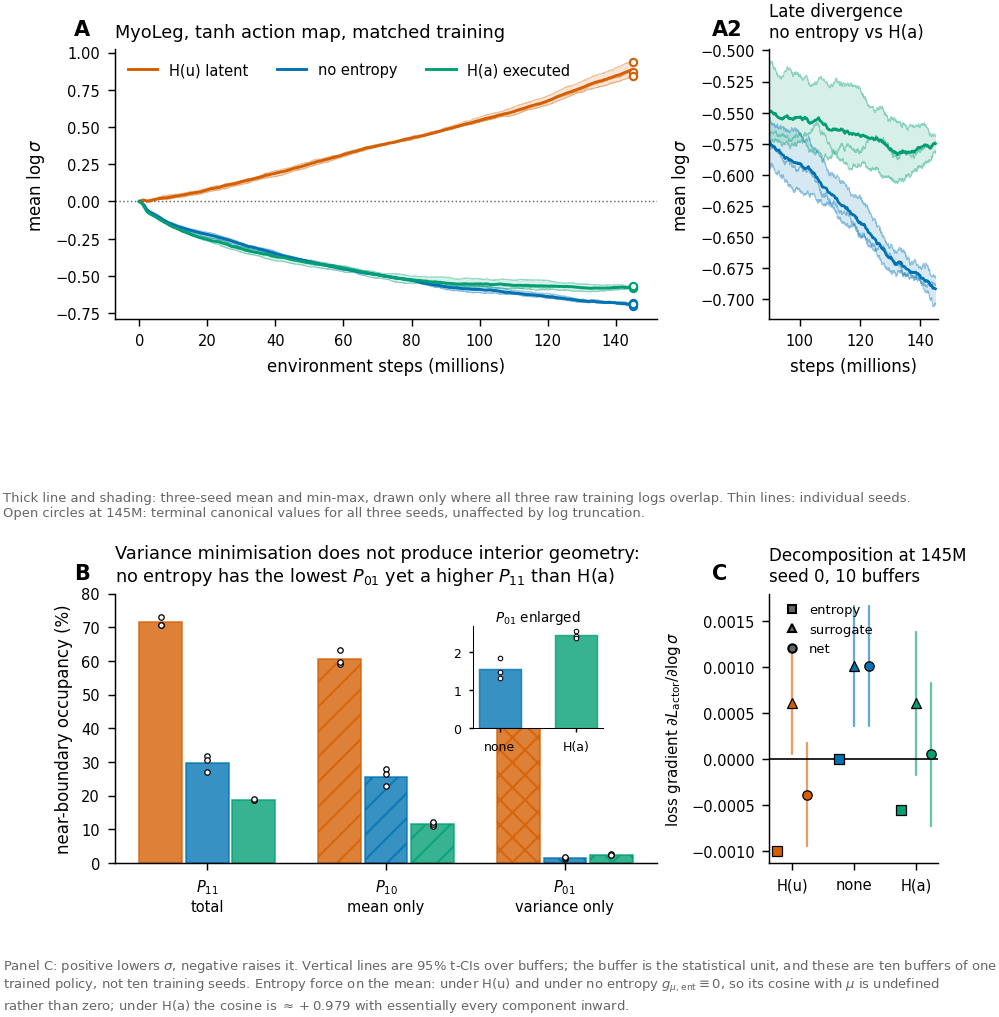}
\caption{Matched MyoLeg tanh experiment. \textbf{A:} Mean $\log\sigma$ during training. Thick lines and shading show the three-seed mean and min--max envelope over the full logged range; thin lines are individual seeds. Open circles are the three terminal 145M canonical values. \textbf{A2:} Separate late-training view of no entropy and \Ha{}. \textbf{B:} Late-training same-state decomposition. No entropy has lower $P_{01}$ than \Ha{} but higher $P_{10}$ and total occupancy. \textbf{C:} Frozen seed-0 gradient decomposition over 10 independent buffers. Positive loss gradient lowers $\sigma$; negative raises it. Under \Hu{} and no entropy, $g_{\mu,\mathrm{ent}}\equiv0$, so its cosine with $\mu$ is undefined. Under \Ha{}, the entropy mean-force cosine is approximately $+0.979$, with essentially every evaluated state--action component receiving an inward contribution.}
\label{fig:myomech}
\end{figure}

Figure~\ref{fig:myomech} follows the three matched tanh conditions over training. Under \Hu{}, mean $\log\sigma$ rises to $+0.8835\pm0.0499$. With no entropy it falls to $-0.6920\pm0.0110$. Under \Ha{} it falls early and reaches $-0.5748\pm0.0069$ at 145M, with little late change. These trajectories are consistent with the direct entropy gradients in Eqs.~\eqref{eq:latentgrad}--\eqref{eq:havar}, while the local frozen-buffer decomposition constrains rather than fully explains the long-horizon training dynamics.

Table~\ref{tab:myo3} shows the late-training geometry. The strongest comparison is between no entropy and \Ha{}. No entropy has the lowest $P_{01}$, 1.53\% versus 2.43\% for \Ha{}. Yet its total occupancy is much higher, 29.76\% versus 18.83\%, because its $P_{10}$ is 25.67\% instead of 11.62\%. At the 5\% reference margin, the condition with the smallest exploration-driven boundary occupancy does not have the smallest total occupancy.

\begin{table}[H]
\centering
\caption{Late-training MyoLeg tanh policies at 145,022,976 steps. Mean $\pm$ sample SD over three training seeds. $P_{11}$ and $P_{10}$ are analytic probabilities (Section~\ref{sec:geometry}) on states visited by the stochastic policy; $P_{01}$ depends only on the state-independent $\sigma$ and involves no state set. The realized closed-loop occupancy agrees with $P_{11}$ within 0.03 percentage points.}
\label{tab:myo3}
\scriptsize
\setlength{\tabcolsep}{2.5pt}
\resizebox{\textwidth}{!}{%
\begin{tabular}{lrrrrrrrr}
\toprule
Condition & $\meanlogsig$ & $\Rstoch$ & $P_{11}$ & $P_{10}$ & $P_{01}$ & $\mathbb{E}|\mu|$ & $\Rdet$ & Det. near bound \\
\midrule
\Hu{} & $+0.8835\pm0.0499$ & $13829\pm635$ & $71.42\pm1.29$ & $60.72\pm2.23$ & $51.80\pm1.48$ & $2.142\pm0.148$ & $899\pm163$ & $50.09\pm1.72$ \\
No entropy & $-0.6920\pm0.0110$ & $\mathbf{16011\pm55}$ & $29.76\pm2.54$ & $25.67\pm2.62$ & $\mathbf{1.53\pm0.27}$ & $1.035\pm0.070$ & $\mathbf{9154\pm2480}$ & $24.77\pm1.71$ \\
\Ha{} & $-0.5748\pm0.0069$ & $15539\pm37$ & $\mathbf{18.83\pm0.28}$ & $\mathbf{11.62\pm0.56}$ & $2.43\pm0.10$ & $\mathbf{0.716\pm0.002}$ & $5001\pm781$ & $\mathbf{10.35\pm1.01}$ \\
\bottomrule
\end{tabular}}
\end{table}

The return ordering is different from the geometry ordering. No entropy has the highest stochastic return in all three matched MyoLeg seeds, while \Ha{} has the most interior mean geometry. Under deterministic execution, \Ha{} again has the lowest near-boundary action rate, but no entropy has the highest return in all three seeds. More interior action geometry is therefore not the same as higher task return, under either stochastic or deterministic execution.

\subsection{The direct gradient on the mean is measurable}
The analytic contrast is exact: \Hu{} has no entropy gradient on $\mu$, whereas \Ha{} does. The frozen MyoLeg buffers reproduce this difference. Under \Ha{}, the entropy contribution satisfies
\begin{equation}
\langle g_{\mu,\mathrm{ent}},\mu\rangle
=5.98\times10^{-4}
\end{equation}
with cosine $+0.979$; essentially all evaluated components receive an inward entropy-gradient contribution. Under \Hu{} and no entropy the entropy gradient with respect to the mean is exactly zero.

The PPO surrogate has no comparably aligned radial direction, but its small radial component is not symmetric either. Its mean gradient has a near-zero cosine with $\mu$ and a componentwise inward fraction near one half, and its projection is condition dependent: $\langle g_{\mu,\mathrm{surr}},\mu\rangle$ averages $-1.06\times10^{-3}$ over the ten no-entropy buffers, with a 95\% interval $[-2.10,-0.02]\times10^{-3}$ that excludes zero, $-4.6\times10^{-4}$ under \Ha{} with an interval spanning zero, and $+1.5\times10^{-4}$ under \Hu{}. Mean extrusion is therefore not a strong universal outward push from the surrogate; it is a weakly outward-leaning, heterogeneous training outcome that latent entropy leaves unconstrained and that the \Ha{} entropy term opposes with a consistently inward projection of comparable magnitude.

For $\log\sigma$, the late-training seed-0 buffer means are informative but local. No entropy has a positive net loss gradient, $+1.015\times10^{-3}$ with a 95\% interval excluding zero, consistent with continued variance reduction. Under \Hu{}, the exact $-10^{-3}$ entropy term offsets a positive surrogate contribution, giving a negative point estimate whose interval spans zero. Under \Ha{}, the entropy term is $-5.52\times10^{-4}$ and the net point estimate is near zero, again with an interval spanning zero. These local measurements support the relative MyoLeg variance regimes without implying that one late-training snapshot explains an entire training trajectory (Appendix~\ref{app:grad}).

\subsection{External replication in Dog-Stand}
\begin{figure}[H]
\centering
\includegraphics[width=\textwidth]{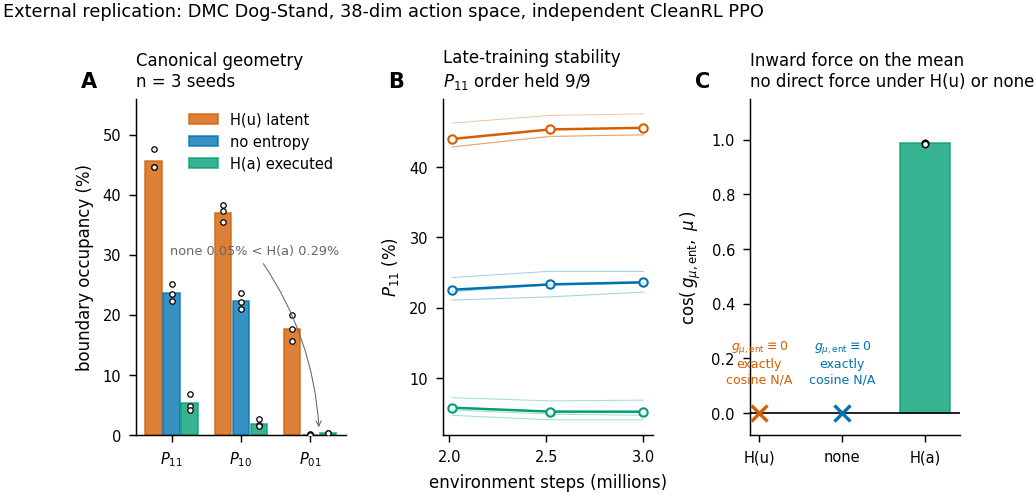}
\caption{External replication on 38-dimensional DMC Dog-Stand with an independent, CleanRL-based PPO implementation. \textbf{A:} Canonical same-state geometry at 3M steps. The variance-only difference between no entropy and \Ha{} is small; the large contrast is in mean placement. \textbf{B:} $P_{11}$ ordering is unchanged at 2.015M, 2.519M, and 3.000M for every training seed (nine of nine checkpoint--seed comparisons). This is late-training stability of the measured geometry, not an asymptotic convergence claim. \textbf{C:} Direct entropy force on the mean. Under \Hu{} and no entropy the entropy mean-gradient vector is exactly zero and the cosine is not defined; crosses mark these zero-vector conditions. Under \Ha{}, the cosine is approximately $+0.987$ across the three seeds.}
\label{fig:dogrep}
\end{figure}

The Dog-Stand result reproduces the central ordering in a different task and a different PPO codebase. At 3M steps, $P_{10}$ is 36.99\% under \Hu{}, 22.25\% with no entropy, and 1.91\% under \Ha{}. Total occupancy follows the same order: 45.57\%, 23.64\%, and 5.26\% (Figure~\ref{fig:dogrep}A). The ordering holds in all three seeds and at all three audited late checkpoints---nine of nine checkpoint--seed comparisons (Figure~\ref{fig:dogrep}B).

The Dog variance trajectories also sharpen the wording of the mechanism. \Hu{} ends at a much higher variance than no entropy, $-0.164$ versus $-1.313$, but its absolute variance decreases late in training. Latent entropy therefore contributes a constant variance-increasing gradient and shifts the learned variance regime upward; it does not force absolute variance to rise in every task. The terminal Dog variance-gradient intervals, measured on frozen buffers, span zero, so the final local balance is not resolved.

In contrast, the inward mean mechanism of \Ha{} is reproduced directly. Across the three Dog seeds, the entropy mean-force cosine is approximately $+0.987$ (Figure~\ref{fig:dogrep}C). The surrogate's projection mirrors MyoLeg: $\langle g_{\mu,\mathrm{surr}},\mu\rangle$ is negative in every no-entropy seed ($-4.3$, $-4.7$, $-3.4\times10^{-3}$; two of the three 95\% intervals exclude zero) and negative but unresolved under \Ha{} ($-0.3$, $-2.2$, $-0.3\times10^{-3}$, all intervals spanning zero), where the entropy term contributes $+3.0$ to $+3.5\times10^{-4}$ inward in every seed. Under \Hu{} and no entropy the entropy gradient with respect to the mean is exactly zero. The main cross-task result is therefore asymmetric: the direct inward mean regularizer is cleanly reproduced, while the detailed long-horizon variance dynamics remain task dependent.

\subsection{State visitation and the 5\% threshold do not explain the primary ordering}
The common-state audit scores each trained Dog policy on every policy's raw visited states, using the evaluated policy's own frozen observation normalizer (Appendix~\ref{app:common}). On the equal-mixture pool,
\begin{equation}
\begin{aligned}
P_{10}:\quad &39.69\%\ (\Hu{}) > 26.99\%\ (\text{none})\\
&> 4.47\%\ (\Ha{}).
\end{aligned}
\end{equation}
The same ordering holds on all six evaluated state sets (the five donor sets plus the equal-mixture pool) and all three training seeds: 18 of 18 state-set--seed comparisons. The largest donor-induced span in seed-averaged $P_{10}$ for any one policy is 6.37 percentage points, while the pooled \Hu{}--\Ha{} separation is 35.23 points. Policy-specific state visitation changes the magnitude but does not explain the ordering.

The same conclusion holds across boundary margins from 1\% to 10\%. For both Dog-Stand and MyoLeg, $P_{10}$ and $P_{11}$ retain the order \Hu{} $>$ no entropy $>$ \Ha{} at 1\%, 2.5\%, 5\%, and 10\% margins, for every training seed---the complete three-condition ordering holds in all 48 task~$\times$~metric~$\times$~margin~$\times$~seed cells. The small variance-only reversal $P_{01}(\text{none})<P_{01}(\Ha{})$ is less robust. In MyoLeg, one no-entropy seed reverses this relation at the 1\% and 2.5\% margins. We therefore use the $P_{01}$ comparison only at the 5\% reference margin and place the main weight on the much larger differences in mean geometry, which are robust across thresholds.

\subsection{Direct mean regularization reproduces centering but not the same operating regime}
The direct mean penalties change an important part of the interpretation. Before running them, we expected the tanh-squared penalty to be too weak once the mean entered the tanh tail, expected both penalties to end near the no-entropy variance, and expected the L2 penalty to trade return for interior geometry. None of these working predictions was supported.

Table~\ref{tab:dog5} shows the canonical five-condition comparison. The tanh-squared penalty reduces $P_{10}$ to 0.09\%, and pre-tanh L2 reduces it to 1.60\%, both below \Ha{} at 1.91\%. Interior means are therefore not unique to the intervention on entropy measurement space. The penalties also end at lower variance than no entropy, even though their raw loss has no derivative with respect to $\log\sigma$. Regularizing the mean alone can indirectly change the long-horizon variance regime through the coupled PPO optimization dynamics.

\begin{table}[H]
\centering
\caption{Dog-Stand canonical comparison at 3,000,320 steps. Mean $\pm$ sample SD over three training seeds. The two direct mean regularizers were added after the primary three-condition experiment as mechanistic controls.}
\label{tab:dog5}
\small
\setlength{\tabcolsep}{3pt}
\resizebox{\textwidth}{!}{%
\begin{tabular}{lrrrrrrr}
\toprule
Condition & $\meanlogsig$ & $\Rstoch$ & $\Rdet$ & $P_{11}$ & $P_{10}$ & $P_{01}$ & $\mathbb{E}|\mu|$ \\
\midrule
\Hu{} & $-0.164\pm0.113$ & $470.5\pm47.2$ & $496.3\pm59.3$ & $45.57\pm1.71$ & $36.99\pm1.46$ & $17.74\pm2.13$ & $1.302\pm0.043$ \\
No entropy & $-1.313\pm0.071$ & $464.1\pm73.5$ & $470.2\pm82.5$ & $23.64\pm1.47$ & $22.25\pm1.26$ & $0.05\pm0.05$ & $0.964\pm0.028$ \\
Tanh$^2$ mean penalty & $-1.459\pm0.048$ & $483.9\pm48.2$ & $455.4\pm52.1$ & $0.20\pm0.15$ & $\mathbf{0.09\pm0.10}$ & $0.05\pm0.04$ & $\mathbf{0.208\pm0.007}$ \\
Pre-tanh L2 & $-1.518\pm0.040$ & $465.1\pm39.4$ & $473.8\pm32.3$ & $2.49\pm0.23$ & $1.60\pm0.25$ & $0.01\pm0.00$ & $0.475\pm0.015$ \\
\Ha{} & $-0.918\pm0.086$ & $\mathbf{594.4\pm34.0}$ & $\mathbf{643.2\pm58.4}$ & $5.26\pm1.46$ & $1.91\pm0.66$ & $0.29\pm0.12$ & $0.489\pm0.030$ \\
\bottomrule
\end{tabular}}
\end{table}

\begin{figure}[H]
\centering
\includegraphics[width=0.88\textwidth]{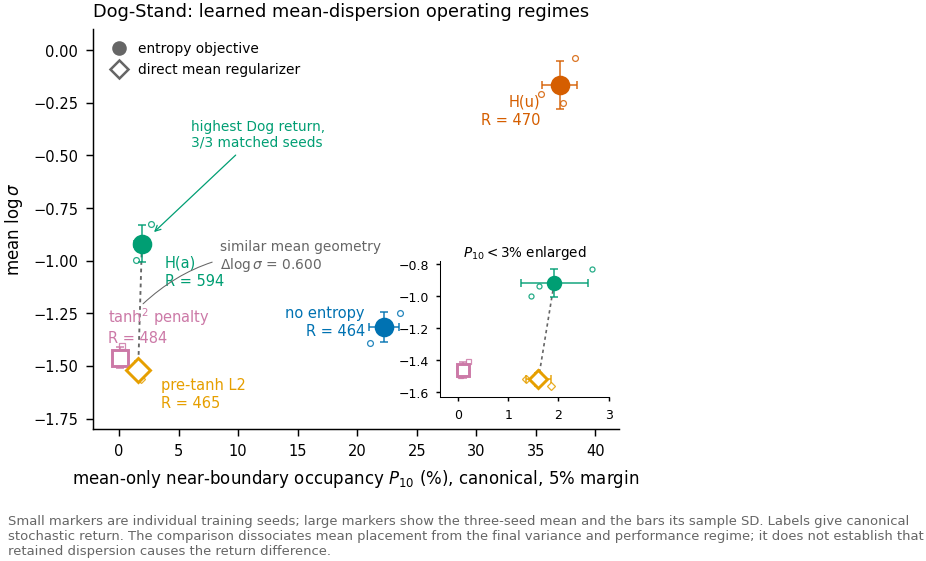}
\caption{Dog-Stand mean--dispersion operating regimes. Small markers are individual training seeds; large markers are three-seed means and bars show sample SD. The horizontal axis is the canonical mean-only cell $P_{10}$ at the 5\% margin; the vertical axis is mean $\log\sigma$. Labels give canonical stochastic return. L2 and \Ha{} have nearly matched mean geometry but substantially different variance and return regimes. The dashed connection highlights this comparison; it does not imply that retained dispersion causes the return difference.}
\label{fig:operating}
\end{figure}

The L2 control is especially informative. Its late-training mean distribution closely matches \Ha{}: $\mathbb{E}|\mu|$ is 0.475 versus 0.489, the 90th/95th/99th percentiles are 0.982/1.178/1.584 versus 1.008/1.211/1.626, and $P_{10}$ is 1.60\% versus 1.91\%. Yet mean $\log\sigma$ differs by 0.600, and stochastic return differs by 129 points. On the pooled set of shared states their $P_{10}$ values remain close, 4.13\% and 4.47\%. Similar mean geometry therefore does not imply a similar variance or performance regime (Figure~\ref{fig:operating}).

In Dog-Stand, \Ha{} has the highest stochastic return among all five conditions in every matched training seed. The three matched-seed differences are all positive, but $n=3$ does not support a population-level performance inference; we therefore treat the return ordering as descriptive sign evidence. This is not a general performance claim: in MyoLeg, no entropy has the highest stochastic return in every matched seed. The consistent cross-task finding is geometric, not a universal return ordering.

\subsection{Boundary occupancy and temporal action structure are related but distinct}
\begin{figure}[H]
\centering
\includegraphics[width=\textwidth]{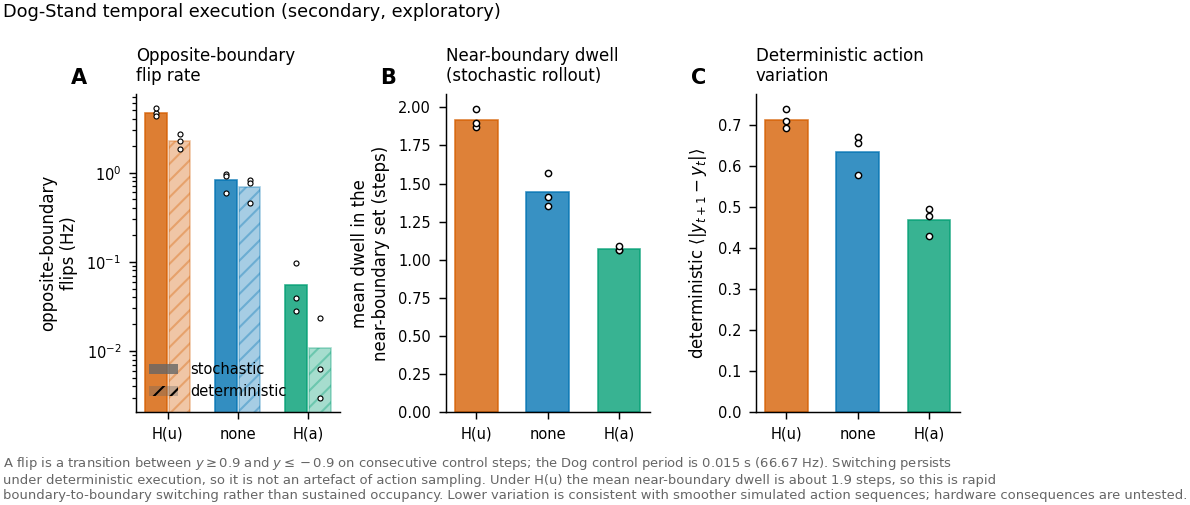}
\caption{Dog-Stand temporal execution structure, reported as secondary exploratory evidence. \textbf{A:} Opposite-boundary flip rate. A flip is a direct transition between $y\ge0.9$ and $y\le-0.9$ on consecutive control steps; the control period is 0.015 s (66.67 Hz). \textbf{B:} Mean dwell time inside the near-boundary set during stochastic rollout. \textbf{C:} Deterministic action variation, $\mathbb{E}|y_{t+1}-y_t|$. Lower variation is consistent with smoother simulated action sequences; hardware consequences are not evaluated.}
\label{fig:temporal}
\end{figure}

Dog-Stand also lets us examine temporal action structure without early terminations. Figure~\ref{fig:temporal} reports three secondary metrics. Under stochastic execution, the opposite-boundary flip rate is 4.711 Hz for \Hu{}, 0.821 Hz for no entropy, and 0.055 Hz for \Ha{}. Under deterministic execution the rates are 2.273, 0.682, and 0.011 Hz. Thus sampling noise alone does not explain the difference. Deterministic mean absolute first difference is 0.7140, 0.6357, and 0.4678, respectively.

These metrics should not be conflated with occupancy. A controller can remain near one bound for a long time and have high occupancy but few flips. Here, however, \Hu{} combines high occupancy with rapid transitions between opposite boundaries. \Ha{} reduces both. These are diagnostics of the present policies, not a comparison with dedicated smooth-control regularizers such as CAPS~\citep{mysore2021caps} or structured-exploration methods such as gSDE~\citep{raffin2022gsde} and colored-noise exploration~\citep{eberhard2023pink}. We add no temporal smoothness loss and use independent Gaussian action sampling in the studied PPO policies. The result is an actuator-level consequence of the learned geometry in simulation, not evidence about wear or hardware safety.

\section{Discussion}
\subsection{What entropy measurement changes}
The central result holds in both studied tasks and does not rest on either one alone. Latent Gaussian entropy changes the variance objective but is blind to the placement of the state-conditioned mean. Executed-action entropy includes the bounded transform and therefore acts on both. In the tanh case, the mean force has a simple direction: it is inward for any nonzero mean. This direct difference appears in the frozen gradients in both MyoLeg and Dog-Stand.

The long-horizon outcome is not determined by the entropy gradient alone. Dog-Stand is the clearest counterexample: \Hu{} has a constant variance-increasing entropy contribution, yet its absolute variance decreases late in training. The correct interpretation is relative. Latent entropy biases the learned policy toward a higher variance regime than matched no-entropy training. Executed entropy adds an inward mean regularizer and a state-dependent variance term. The rest of PPO---the surrogate, visited states, optimizer history, and shared gradient clipping---still participates in the final operating point.

\subsection{Executed entropy is not the only way to center the mean}
The controls with direct mean penalties prevent an overly strong conclusion. A simple explicit penalty can produce an interior policy mean, and the tanh-squared control has lower Dog $P_{10}$ than \Ha{}. Executed entropy is therefore not uniquely capable of preventing mean extrusion. The two penalties are mechanism controls, not claims of a new regularization family. CAPS, for example, explicitly regularizes smoothness of the state-to-action mapping and temporal actions~\citep{mysore2021caps}; our controls penalize only the current state's policy mean and do not target action-rate smoothness.

What distinguishes \Ha{} empirically is the coupled regime it produces. The pre-tanh L2 policy provides an unusually clean comparison because its late-training mean distribution is almost the same as \Ha{} under both on-policy and shared-state evaluation. The two policies nonetheless end with substantially different dispersion and Dog return. This dissociates mean geometry from the final variance/performance regime. It does not show that higher retained variance causes higher return; the two training objectives also change the entire optimization trajectory.

\subsection{Return does not identify policy geometry}
The two tasks make this point in complementary ways. In MyoLeg, no entropy gives the highest return among the three tanh conditions, whereas \Ha{} produces the most interior geometry. In Dog-Stand, \Ha{} gives the highest return among all five conditions. The geometric ordering is consistent across tasks, whereas the return ordering is not. A method selected only by return would therefore lead to different conclusions about action geometry on the two tasks.

The gap between stochastic and deterministic execution is also separable from boundary geometry. In MyoLeg, deterministic return is much lower than stochastic return under all three tanh conditions. In Dog-Stand, the deterministic-to-stochastic return ratio is at or above one for four of the five conditions, including the highly boundary-oriented \Hu{} policy; only the tanh-squared control falls slightly below one (Table~\ref{tab:dog5}). Boundary-heavy geometry is not sufficient to cause a performance gap between stochastic and deterministic execution.

\subsection{Relation to SAC and transformed distributions}
The executed-action entropy used here is standard change-of-variables probability. SAC's tanh-squashed Gaussian already evaluates the transformed log density, including the Jacobian~\citep{haarnoja2018sac}, and it inherits that treatment from the maximum-entropy formulation it instantiates~\citep{ziebart2008maxent,haarnoja2017softq}. The novelty is not the formula. The contribution is the mechanistic diagnosis of what is lost when PPO measures entropy only in the latent Gaussian. The entropy term can then increase variance while exerting no direct constraint on where the mean sits relative to a bounded action map.

This distinction also explains why ``bounded Gaussian mismatch'' is too broad a description of the result. Hard clipping is unnecessary for the high-variance latent regime, and direct mean penalties can repair mean placement without using an entropy objective. The relevant design choice is how the training objective treats both mean placement and dispersion before the bounded execution map.

\subsection{Practical diagnostic}
The four-cell decomposition gives a simple way to diagnose bounded-policy behavior. Report $P_{10}$ and $P_{01}$ in addition to return and total near-boundary occupancy. If $P_{01}$ is large while $P_{10}$ is small, the problem is mainly dispersion and variance tuning may be sufficient. If $P_{10}$ is already large, reducing variance alone cannot produce an interior mean; the policy needs either a geometric term in the executed action space or an explicit mean regularizer.

Whether \Ha{} is worth using depends on the objective. If task return is the only target, neither task supports a universal recommendation. If near-boundary mean placement, rapid boundary switching, or simulated action variation matter, \Ha{} is a principled option because it regularizes the same bounded geometry used for execution. Explicit mean penalties are another option and may produce a different variance regime.

\subsection{Scope}
The experiments identify a reproducible regime, not the minimal conditions required for it. Both tasks use PPO, a diagonal Gaussian latent policy with a state-independent learned $\log\sigma$, bounded actions, and MuJoCo-family simulation. State-dependent standard deviations are a common alternative implementation choice~\citep{andrychowicz2021what}, and gSDE goes further by making exploration state dependent and temporally persistent~\citep{raffin2022gsde}; neither is tested here. High action dimensionality may amplify the phenomenon, but no dimension sweep isolates that factor. Neither PPO specificity nor Gaussian-policy specificity is tested. The entropy coefficient is fixed at $10^{-3}$ in the primary entropy conditions; the analytic gradient direction is known, but the relation between effect size and coefficient is not measured.

The Dog conclusions are also tied to the attained performance regime. Canonical stochastic returns span roughly 464--594 on 1000-step episodes. Whether substantially stronger training, a different optimizer schedule, or another algorithm preserves the same geometry remains open. No hardware experiment is performed. These scope limits do not change the matched comparisons reported here; they define the next questions rather than additional experiments required for the present mechanism claim.

\section{Conclusion}
PPO with a diagonal Gaussian policy and bounded actions has, in the settings studied here, a policy-geometry problem that return alone can hide. In MyoLeg, a boundary-dominated Gaussian policy is produced jointly by mean extrusion and dispersion; the mean by itself is already sufficient for most of the occupancy. Measuring entropy in the latent Gaussian raises the relative variance regime without constraining mean placement. Measuring entropy after the bounded transform adds a direct inward mean regularizer and changes the variance objective. The resulting ordering in mean geometry is reproduced in a 38-dimensional Dog-Stand task, survives evaluation on shared states, and is unchanged across boundary margins from 1\% to 10\%.

Executed entropy is not the only way to obtain interior means. Direct mean penalties can match or exceed its centering. Their late-training variance and performance regimes can still differ substantially, showing that mean placement, dispersion, and task return are distinct properties of a learned controller. For bounded-action PPO, all three should be measured explicitly.

\bibliographystyle{tmlr}
\bibliography{references}

@article{vandenbogert2025,
  title={Strange Effects of Activation Dynamics on Musculoskeletal Trajectory Optimization},
  author={van den Bogert, Antonie J.},
  journal={bioRxiv},
  year={2025},
  doi={10.1101/2025.01.30.635759},
  note={Preprint; not peer reviewed}
}

@inproceedings{caggiano2022myosuite,
  title={MyoSuite: A Contact-rich Simulation Suite for Musculoskeletal Motor Control},
  author={Caggiano, Vittorio and Wang, Huawei and Durandau, Guillaume and Sartori, Massimo and Kumar, Vikash},
  booktitle={Proceedings of the 4th Annual Learning for Dynamics and Control Conference},
  series={Proceedings of Machine Learning Research},
  volume={168},
  pages={492--507},
  year={2022}
}

@inproceedings{caggiano2024myochallenge,
  title={{MyoChallenge} 2024: A New Benchmark for Physiological Dexterity and Agility in Bionic Humans},
  author={Wang, Cheryl and Tan, Chun Kwang and Hodossy, Balint and Lyu, Eric and Schumacher, Pierre and Heald, James and Biegun, Kai and Hromadka, Samo and Sahani, Maneesh and Park, Gunwoo and Shin, Beomsoo and Park, Jonghyun and Koo, Seungbum and Zuo, Chenhui and Ma, Chengtian and Sui, Yanan and Hansen, Nicklas and Tao, Stone and Gao, Yuan and Su, Hao and Song, Seungmoon and Gionfrida, Letizia and Sartori, Massimo and Durandau, Guillaume and Kumar, Vikash and Caggiano, Vittorio},
  booktitle={Advances in Neural Information Processing Systems},
  volume={38},
  note={Datasets and Benchmarks Track},
  year={2025}
}

@incollection{kidzinski2018learning,
  title={Learning to Run Challenge Solutions: Adapting Reinforcement Learning Methods for Neuromusculoskeletal Environments},
  author={Kidzi{\'n}ski, {\L}ukasz and Mohanty, Sharada Prasanna and Ong, Carmichael F. and Huang, Zhewei and Zhou, Shuchang and Pechenko, Anton and others},
  booktitle={The NIPS '17 Competition: Building Intelligent Systems},
  pages={121--153},
  publisher={Springer},
  year={2018},
  doi={10.1007/978-3-319-94042-7_7}
}

@article{schulman2017ppo,
  title={Proximal Policy Optimization Algorithms},
  author={Schulman, John and Wolski, Filip and Dhariwal, Prafulla and Radford, Alec and Klimov, Oleg},
  journal={arXiv preprint arXiv:1707.06347},
  year={2017}
}

@inproceedings{andrychowicz2021what,
  title={What Matters for On-Policy Deep Actor-Critic Methods? A Large-Scale Study},
  author={Andrychowicz, Marcin and Raichuk, Anton and Sta{\'n}czyk, Piotr and Orsini, Manu and Girgin, Sertan and Marinier, Rapha{\"e}l and Hussenot, Leonard and Geist, Matthieu and Pietquin, Olivier and Michalski, Marcin and Gelly, Sylvain and Bachem, Olivier},
  booktitle={International Conference on Learning Representations},
  year={2021}
}

@inproceedings{schumacher2022deprl,
  title={DEP-RL: Embodied Exploration for Reinforcement Learning in Overactuated and Musculoskeletal Systems},
  author={Schumacher, Pierre and Haeufle, Daniel F. B. and B{\"u}chler, Dieter and Schmitt, Syn and Martius, Georg},
  booktitle={International Conference on Learning Representations},
  year={2023}
}

@article{lee2019scalable,
  title={Scalable Muscle-Actuated Human Simulation and Control},
  author={Lee, Seunghwan and Park, Moonseok and Lee, Kyoungmin and Lee, Jehee},
  journal={ACM Transactions on Graphics},
  volume={38},
  number={4},
  pages={73:1--73:13},
  year={2019},
  doi={10.1145/3306346.3322972}
}

@article{anderson2001dynamic,
  title={Dynamic Optimization of Human Walking},
  author={Anderson, Frank C. and Pandy, Marcus G.},
  journal={Journal of Biomechanical Engineering},
  volume={123},
  number={5},
  pages={381--390},
  year={2001},
  doi={10.1115/1.1392310}
}

@article{raffin2021sb3,
  title={Stable-Baselines3: Reliable Reinforcement Learning Implementations},
  author={Raffin, Antonin and Hill, Ashley and Gleave, Adam and Kanervisto, Anssi and Ernestus, Maximilian and Dormann, Noah},
  journal={Journal of Machine Learning Research},
  volume={22},
  number={268},
  pages={1--8},
  year={2021}
}

@inproceedings{todorov2012mujoco,
  title={MuJoCo: A Physics Engine for Model-Based Control},
  author={Todorov, Emanuel and Erez, Tom and Tassa, Yuval},
  booktitle={IEEE/RSJ International Conference on Intelligent Robots and Systems},
  pages={5026--5033},
  year={2012},
  doi={10.1109/IROS.2012.6386109}
}

@inproceedings{chou2017beta,
  title={Improving Stochastic Policy Gradients in Continuous Control with Deep Reinforcement Learning using the {Beta} Distribution},
  author={Chou, Po-Wei and Maturana, Daniel and Scherer, Sebastian},
  booktitle={Proceedings of the 34th International Conference on Machine Learning},
  series={Proceedings of Machine Learning Research},
  volume={70},
  pages={834--843},
  year={2017}
}

@inproceedings{petrazzini2021beta,
  title={Proximal Policy Optimization with Continuous Bounded Action Space via the {Beta} Distribution},
  author={Petrazzini, Irving G. B. and Antonelo, Eric A.},
  booktitle={2021 IEEE Symposium Series on Computational Intelligence},
  pages={1--8},
  year={2021},
  doi={10.1109/SSCI50451.2021.9660123}
}

@inproceedings{fujita2018clipped,
  title={Clipped Action Policy Gradient},
  author={Fujita, Yasuhiro and Maeda, Shin-ichi},
  booktitle={Proceedings of the 35th International Conference on Machine Learning},
  series={Proceedings of Machine Learning Research},
  volume={80},
  pages={1597--1606},
  year={2018}
}

@inproceedings{haarnoja2018sac,
  title={Soft Actor-Critic: Off-Policy Maximum Entropy Deep Reinforcement Learning with a Stochastic Actor},
  author={Haarnoja, Tuomas and Zhou, Aurick and Abbeel, Pieter and Levine, Sergey},
  booktitle={Proceedings of the 35th International Conference on Machine Learning},
  series={Proceedings of Machine Learning Research},
  volume={80},
  pages={1861--1870},
  year={2018}
}

@inproceedings{chiappa2023lattice,
  title={Latent Exploration for Reinforcement Learning},
  author={Chiappa, Alberto Silvio and Marin Vargas, Alessandro and Huang, Ann and Mathis, Alexander},
  booktitle={Advances in Neural Information Processing Systems},
  volume={36},
  year={2023}
}

@inproceedings{seyde2021bangbang,
  title={Is Bang-Bang Control All You Need? Solving Continuous Control with Bernoulli Policies},
  author={Seyde, Tim and Gilitschenski, Igor and Schwarting, Wilko and Stellato, Bartolomeo and Riedmiller, Martin and Wulfmeier, Markus and Rus, Daniela},
  booktitle={Advances in Neural Information Processing Systems},
  volume={34},
  pages={27209--27221},
  year={2021}
}

@article{huang2022cleanrl,
  title={CleanRL: High-Quality Single-File Implementations of Deep Reinforcement Learning Algorithms},
  author={Huang, Shengyi and Dossa, Rousslan Fernand Julien and Ye, Chang and Braga, Jeff and Chakraborty, Dipam and Mehta, Kinal and Ara{\'u}jo, Jo{\~a}o G. M.},
  journal={Journal of Machine Learning Research},
  volume={23},
  number={274},
  pages={1--18},
  year={2022}
}

@article{tassa2020dmcontrol,
  title={dm\_control: Software and Tasks for Continuous Control},
  author={Tunyasuvunakool, Saran and Muldal, Alistair and Doron, Yotam and Liu, Siqi and Bohez, Steven and Merel, Josh and Erez, Tom and Lillicrap, Timothy and Heess, Nicolas and Tassa, Yuval},
  journal={Software Impacts},
  volume={6},
  pages={100022},
  year={2020},
  doi={10.1016/j.simpa.2020.100022}
}

@article{zhou2026gaussian,
  title={Influence of Gaussian Distribution on Performance Metrics in Continuous Reinforcement Learning},
  author={Zhou, Ruikai and Zhong, Taihong and Zhu, Wenbo and Han, Shuai and L{\"u}, Shuai},
  journal={Information Processing \& Management},
  volume={63},
  number={2},
  pages={104428},
  year={2026},
  doi={10.1016/j.ipm.2025.104428}
}

@inproceedings{mysore2021caps,
  title={Regularizing Action Policies for Smooth Control with Reinforcement Learning},
  author={Mysore, Siddharth and Mabsout, Bassel and Mancuso, Renato and Saenko, Kate},
  booktitle={IEEE International Conference on Robotics and Automation},
  pages={1810--1816},
  year={2021},
  doi={10.1109/ICRA48506.2021.9561138}
}

@inproceedings{raffin2022gsde,
  title={Smooth Exploration for Robotic Reinforcement Learning},
  author={Raffin, Antonin and Kober, Jens and Stulp, Freek},
  booktitle={Proceedings of the 5th Conference on Robot Learning},
  series={Proceedings of Machine Learning Research},
  volume={164},
  pages={1634--1644},
  year={2022}
}

@inproceedings{eberhard2023pink,
  title={Pink Noise Is All You Need: Colored Noise Exploration in Deep Reinforcement Learning},
  author={Eberhard, Onno and Hollenstein, Jakob and Pinneri, Cristina and Martius, Georg},
  booktitle={International Conference on Learning Representations},
  year={2023}
}

@misc{cmumocap,
  year={2003},
  author={{Carnegie Mellon University Graphics Lab}},
  title={{CMU Graphics Lab Motion Capture Database}},
  howpublished={\url{https://mocap.cs.cmu.edu/}},
  note={Subject 2, trial 1 used as \texttt{02\_01.c3d}; accessed Aug. 17, 2026}
}

@inproceedings{schulman2016gae,
  title={High-Dimensional Continuous Control Using Generalized Advantage Estimation},
  author={Schulman, John and Moritz, Philipp and Levine, Sergey and Jordan, Michael I. and Abbeel, Pieter},
  booktitle={International Conference on Learning Representations},
  year={2016}
}

@article{williams1991entropy,
  title={Function Optimization Using Connectionist Reinforcement Learning Algorithms},
  author={Williams, Ronald J. and Peng, Jing},
  journal={Connection Science},
  volume={3},
  number={3},
  pages={241--268},
  year={1991},
  doi={10.1080/09540099108946587}
}

@inproceedings{mnih2016a3c,
  title={Asynchronous Methods for Deep Reinforcement Learning},
  author={Mnih, Volodymyr and Badia, Adri{\`a} Puigdom{\`e}nech and Mirza, Mehdi and Graves, Alex and Lillicrap, Timothy P. and Harley, Tim and Silver, David and Kavukcuoglu, Koray},
  booktitle={Proceedings of the 33rd International Conference on Machine Learning},
  series={Proceedings of Machine Learning Research},
  volume={48},
  pages={1928--1937},
  year={2016}
}

@inproceedings{ziebart2008maxent,
  title={Maximum Entropy Inverse Reinforcement Learning},
  author={Ziebart, Brian D. and Maas, Andrew L. and Bagnell, J. Andrew and Dey, Anind K.},
  booktitle={Proceedings of the 23rd AAAI Conference on Artificial Intelligence},
  pages={1433--1438},
  year={2008}
}

@inproceedings{haarnoja2017softq,
  title={Reinforcement Learning with Deep Energy-Based Policies},
  author={Haarnoja, Tuomas and Tang, Haoran and Abbeel, Pieter and Levine, Sergey},
  booktitle={Proceedings of the 34th International Conference on Machine Learning},
  series={Proceedings of Machine Learning Research},
  volume={70},
  pages={1352--1361},
  year={2017}
}

@inproceedings{engstrom2020implementation,
  title={Implementation Matters in Deep Policy Gradients: A Case Study on {PPO} and {TRPO}},
  author={Engstrom, Logan and Ilyas, Andrew and Santurkar, Shibani and Tsipras, Dimitris and Janoos, Firdaus and Rudolph, Larry and Madry, Aleksander},
  booktitle={International Conference on Learning Representations},
  year={2020}
}

@inproceedings{kingma2015adam,
  title={{Adam}: A Method for Stochastic Optimization},
  author={Kingma, Diederik P. and Ba, Jimmy},
  booktitle={International Conference on Learning Representations},
  year={2015}
}

@article{towers2024gymnasium,
  title={Gymnasium: A Standard Interface for Reinforcement Learning Environments},
  author={Towers, Mark and Kwiatkowski, Ariel and Terry, Jordan and Balis, John U. and De Cola, Gianluca and Deleu, Tristan and Goul{\~a}o, Manuel and Kallinteris, Andreas and Krimmel, Markus and KG, Arjun and Perez-Vicente, Rodrigo and Pierr{\'e}, Andrea and Schulhoff, Sander and Tai, Jun Jet and Tan, Hannah and Younis, Omar G.},
  journal={arXiv preprint arXiv:2407.17032},
  year={2024}
}

@misc{shimmy2023,
  author={Tai, Jun Jet and Towers, Mark and Tower, Elliot},
  title={Shimmy: {Gymnasium} and {PettingZoo} Wrappers for Commonly Used Environments},
  year={2023},
  publisher={Zenodo},
  doi={10.5281/zenodo.8140744},
  howpublished={\url{https://github.com/Farama-Foundation/Shimmy}}
}

@inproceedings{freeman2021brax,
  title     = {Brax -- A Differentiable Physics Engine for Large Scale Rigid Body Simulation},
  author    = {Freeman, C. Daniel and Frey, Erik and Raichuk, Anton and Girgin, Sertan and Mordatch, Igor and Bachem, Olivier},
  booktitle = {Advances in Neural Information Processing Systems (NeurIPS), Datasets and Benchmarks Track},
  year      = {2021},
  note      = {arXiv:2106.13281}
}

@inproceedings{eisenach2019marginal,
  title     = {Marginal Policy Gradients: A Unified Family of Estimators for Bounded Action Spaces with Applications},
  author    = {Eisenach, Carson and Yang, Haichuan and Liu, Ji and Liu, Han},
  booktitle = {International Conference on Learning Representations (ICLR)},
  year      = {2019}
}

@article{lin2025gac,
  title   = {Beyond Distributions: Geometric Action Control for Continuous Reinforcement Learning},
  author  = {Lin, Zhihao},
  journal = {arXiv preprint arXiv:2511.08234},
  year    = {2025}
}

\appendix

\section{Boundary Margin Sensitivity}
\label{app:margin}
Table~\ref{tab:margins} reports the seed-mean $P_{11}/P_{10}$ values used for the margin-robustness statement. For each checkpoint, one fresh 30-episode on-policy state set is collected and all four margins are evaluated analytically on those same states; no actions are resampled for the boundary probabilities. Both sweeps use the same NumPy action-noise convention as their frozen canonical evaluators, so each checkpoint's state set coincides with its canonical one and the 5\% cells reproduce Tables~\ref{tab:myo3} and \ref{tab:dog5} exactly.

\begin{table}[H]
\centering
\caption{Boundary-margin sensitivity. Entries are $P_{11}/P_{10}$ in percent, averaged over three training seeds. The primary ordering \Hu{} $>$ no entropy $>$ \Ha{} holds for both quantities in every seed at every margin. Both 5\% rows reproduce their frozen canonical values (Tables~\ref{tab:myo3} and \ref{tab:dog5}) exactly.}
\label{tab:margins}
\small
\begin{tabular}{llccc}
\toprule
Task & Margin & \Hu{} & No entropy & \Ha{} \\
\midrule
Dog & 1\% & 24.47 / 15.83 & 6.63 / 5.88 & 0.28 / 0.05 \\
Dog & 2.5\% & 35.34 / 26.31 & 14.18 / 13.02 & 1.63 / 0.44 \\
Dog & 5\% & 45.57 / 36.99 & 23.64 / 22.25 & 5.26 / 1.91 \\
Dog & 10\% & 57.77 / 50.46 & 37.58 / 36.10 & 14.69 / 7.44 \\
\midrule
MyoLeg & 1\% & 56.40 / 40.46 & 10.30 / 7.69 & 2.77 / 0.82 \\
MyoLeg & 2.5\% & 64.70 / 51.62 & 19.51 / 15.81 & 9.28 / 4.75 \\
MyoLeg & 5\% & 71.42 / 60.72 & 29.76 / 25.67 & 18.83 / 11.62 \\
MyoLeg & 10\% & 78.61 / 70.54 & 43.55 / 39.42 & 33.50 / 23.19 \\
\bottomrule
\end{tabular}
\end{table}

The 5\% regression check uses a 0.05-percentage-point tolerance as a guard against state-timing, normalization, threshold, and cell-definition errors; in the released runs the reproduction is exact. Table~\ref{tab:myo3} remains the source of record for the primary 5\% MyoLeg values, and Table~\ref{tab:margins} for the four-margin sweep.

The variance-only comparison between no entropy and \Ha{} is not threshold invariant in MyoLeg. At 1\% and 2.5\%, no-entropy seed 2 has $P_{01}=0.2610\%$ and 0.7309\%, above \Ha{} at 0.1204\% and 0.6857\%. At the 5\% reference margin the same seed has 1.8284\% versus 2.3493\%, and at 10\% it has 5.2824\% versus 7.5610\%. This small comparison is therefore not used as a margin-robust headline result.

\section{Common-State Dog Audit}
\label{app:common}
Table~\ref{tab:common} reports $P_{10}$ when every trained Dog policy is evaluated on every policy's raw donor states. Each evaluated policy applies its own frozen observation normalizer. The equal-mixture pool contains an equal number of states from each donor condition.

\begin{table}[H]
\centering
\caption{Dog common-state $P_{10}$ (\%), donor state set in rows and evaluated policy in columns, mean over three training seeds.}
\label{tab:common}
\small
\begin{tabular}{lrrrrr}
\toprule
Donor & \Hu{} & No entropy & \Ha{} & Tanh$^2$ penalty & Pre-tanh L2 \\
\midrule
\Hu{} & 37.05 & 28.31 & 4.56 & 0.59 & 5.30 \\
No entropy & 38.00 & 22.26 & 5.15 & 0.96 & 4.23 \\
\Ha{} & 42.06 & 28.62 & 1.88 & 0.57 & 4.52 \\
Tanh$^2$ penalty & 42.46 & 27.31 & 5.29 & 0.09 & 5.13 \\
Pre-tanh L2 & 38.89 & 28.52 & 5.38 & 0.59 & 1.60 \\
Equal-mixture pool & 39.69 & 26.99 & 4.47 & 0.56 & 4.13 \\
\bottomrule
\end{tabular}
\end{table}

The primary ordering \Hu{} $>$ no entropy $>$ \Ha{} holds on all six evaluated state sets (the five donor sets plus the equal-mixture pool) in every seed: 18 of 18 state-set--seed comparisons. At the seed-averaged level all five policies have their lowest $P_{10}$ on their own donor distribution; this tendency holds in 14 of 15 individual policy--seed comparisons. We treat this as descriptive evidence of policy--state-distribution co-adaptation, not as a causal result.

\section{Gradient Details}
\label{app:grad}
Table~\ref{tab:grad-detail} reports the local actor-loss gradient decomposition for the late-training MyoLeg seed-0 checkpoint of each entropy condition. Each entry is the mean per-dimension gradient with respect to $\log\sigma$ over 10 independently collected frozen-policy buffers.
\begin{table}[H]
\centering
\caption{MyoLeg frozen-buffer $\log\sigma$ gradient decomposition, 10 independent buffers per condition. Intervals are 95\% $t$-intervals over buffers. Positive loss gradient lowers $\sigma$.}
\label{tab:grad-detail}
\scriptsize
\setlength{\tabcolsep}{2.3pt}
\begin{tabular}{lrrr}
\toprule
Condition & Entropy & Surrogate & Net \\
\midrule
No entropy & $0$ & $+1.015\!\times\!10^{-3}$ & $+1.015\!\times\!10^{-3}$ \\
 & & $[0.363,1.67]\!\times\!10^{-3}$ & same \\
\Hu{} & $-1.000\!\times\!10^{-3}$ & $+0.610\!\times\!10^{-3}$ & $-0.390\!\times\!10^{-3}$ \\
 & exact & $[0.052,1.17]\!\times\!10^{-3}$ & $[-0.949,0.168]\!\times\!10^{-3}$ \\
\Ha{} & $-0.552\!\times\!10^{-3}$ & $+0.604\!\times\!10^{-3}$ & $+0.052\!\times\!10^{-3}$ \\
 & & $[-0.175,1.38]\!\times\!10^{-3}$ & $[-0.727,0.831]\!\times\!10^{-3}$ \\
\bottomrule
\end{tabular}
\end{table}
Only the net interval under no entropy excludes zero. The local decomposition is therefore used to constrain the late-training MyoLeg mechanism, not as a complete causal explanation of the full trajectories.

\section{Training Configuration and Executed-Entropy Estimator}
\label{app:config}
Tables~\ref{tab:hyperparams} and~\ref{tab:networks} give the verified training configuration and network architecture used by the two primary implementations. The MyoLeg source does not override several Stable-Baselines3 PPO fields. For those fields, Table~\ref{tab:hyperparams} reports the resolved values read back from the trained checkpoints by the release manifest, not defaults assumed from library documentation; the manifest also records the environment lockfile.

\begin{table}[H]
\centering
\caption{Training configuration for the primary tanh experiments. The Dog controls with direct mean penalties use the Dog column with $c_{\mathrm{ent}}=0$ plus the stated penalty.}
\label{tab:hyperparams}
\scriptsize
\setlength{\tabcolsep}{4.2pt}
\begin{tabular}{p{0.22\textwidth}p{0.35\textwidth}p{0.35\textwidth}}
\toprule
Setting & MyoLeg / Stable-Baselines3 PPO & Dog-Stand / CleanRL PPO \\
\midrule
Observation / action dimensions & 101 / 80 latent muscle actions & 223 / 38 latent actions \\
Parallel environments & 12, \texttt{SubprocVecEnv} & 1, synchronous vector environment \\
Rollout length & 2048 per environment & 2048 \\
Rollout batch & 24,576 transitions & 2,048 transitions \\
Minibatch size & 128 & 64 (32 minibatches) \\
Update epochs & 10 & 10 \\
Optimizer & Adam, Stable-Baselines3 implementation & Adam, $\epsilon=10^{-5}$ \\
Learning rate & $10^{-4}$, constant & $3\times10^{-4}$, linearly annealed over the prescribed run \\
Discount $\gamma$ & 0.99 & 0.99 \\
GAE $\lambda$ & 0.99 & 0.95 \\
PPO clip coefficient & 0.2 & 0.2 \\
Value-loss coefficient & 0.5 & 0.5 \\
Value clipping & disabled (\texttt{clip\_range\_vf}$=$\texttt{None}) & enabled with clip coefficient 0.2 \\
Advantage normalization & enabled & enabled \\
State-dependent exploration / gSDE & not used; diagonal state-independent $\log\sigma$ & not used; diagonal state-independent $\log\sigma$ \\
Target KL & none (\texttt{target\_kl}$=$\texttt{None}) & none \\
Maximum gradient norm & 0.5 & 0.5 \\
Initial $\log\sigma$ & 0 & 0 \\
Standard deviation & learned state-independent vector & learned state-independent vector \\
Observation normalization & none & running mean/variance, then clipped to $[-10,10]$; frozen during canonical evaluation \\
Reward normalization & none & running reward normalization and $[-10,10]$ clipping during training; canonical evaluation reports raw environment return \\
Action mapping & exposed latent box $[-10,10]^{80}$, with measured clip incidence and its inertness reported in Section~\ref{sec:methods} ($\tanh 10 = 1-4.1\times10^{-9}$); $a=(\tanh u+1)/2$ & \emph{unbounded} exposed latent box; no clip of any kind precedes the transform; $y=\tanh u$, then affine map to physical bounds \\
Entropy coefficient & $10^{-3}$ for \Hu{}/\Ha{}, $0$ for no entropy & $10^{-3}$ for \Hu{}/\Ha{}, $0$ for no entropy \\
Executed-entropy samples & $K=4$, antithetic reparameterized samples & $K=4$, antithetic reparameterized samples \\
Training seeds & 0, 1, 2 & 0, 1, 2 \\
Prescribed environment steps & 145,022,976 & 3,000,320 \\
\bottomrule
\end{tabular}
\end{table}

\begin{table}[H]
\centering
\caption{Policy and value network architecture. Both implementations use separate actor and critic MLPs and a state-independent learnable $\log\sigma$ vector.}
\label{tab:networks}
\scriptsize
\begin{tabular}{p{0.20\textwidth}p{0.36\textwidth}p{0.36\textwidth}}
\toprule
Component & MyoLeg & Dog-Stand \\
\midrule
Actor mean & $101\rightarrow512\rightarrow512\rightarrow80$, ReLU hidden layers & $223\rightarrow64\rightarrow64\rightarrow38$, tanh hidden layers \\
Critic & $101\rightarrow512\rightarrow512\rightarrow1$, ReLU hidden layers & $223\rightarrow64\rightarrow64\rightarrow1$, tanh hidden layers \\
Actor dispersion & 80 learned $\log\sigma_i$ parameters, initialized to zero & 38 learned $\log\sigma_i$ parameters, initialized to zero \\
Initialization & Stable-Baselines3 \texttt{MlpPolicy} initialization & orthogonal; hidden gain $\sqrt{2}$, actor-output gain 0.01, critic-output gain 1.0, zero biases \\
Trainable parameters & 671,393 & 39,565 \\
\bottomrule
\end{tabular}
\end{table}

\subsection{Four-sample antithetic estimator for \texorpdfstring{$H(a)$}{H(a)}}
For a minibatch of states, the executed-entropy implementation uses the following pathwise estimator. The MyoLeg affine map to $[0,1]$ is included in the Jacobian; for Dog, the later affine physical-action scaling is constant in the policy parameters and can be omitted from gradients.

\begin{enumerate}[leftmargin=1.7em,label=\arabic*.]
\item Compute $\sigma=\exp(\log\sigma)$ and the exact latent entropy
$H_u=\sum_i[\log\sigma_i+\tfrac12\log(2\pi e)]$.
\item Draw two independent $\epsilon_1,\epsilon_2\sim\mathcal{N}(0,I)$ and form the four-sample antithetic set $\{\epsilon_1,-\epsilon_1,\epsilon_2,-\epsilon_2\}$.
\item For each $\epsilon_k$, form the reparameterized latent action $u_k=\mu+\sigma\odot\epsilon_k$.
\item Evaluate the stable componentwise log-Jacobian. For $a=(\tanh u+1)/2$,
\begin{equation}
\ell_{[0,1]}(u)=\log 2-2u-2\,\operatorname{softplus}(-2u).
\end{equation}
For the normalized Dog action $y=\tanh u$,
\begin{equation}
\ell_{[-1,1]}(u)=\log 4-2u-2\,\operatorname{softplus}(-2u).
\end{equation}
\item Return
\begin{equation}
\widehat H(a)=H_u+\frac{1}{4}\sum_{k=1}^{4}\sum_i \ell(u_{k,i}),
\end{equation}
and add $-c_{\mathrm{ent}}$ times the minibatch mean of this quantity to the actor loss.
\end{enumerate}

The Gaussian entropy term is analytic; Monte Carlo noise enters only through the expected log-Jacobian. Reparameterization provides pathwise gradients in both $\mu$ and $\log\sigma$. The antithetic construction forces the sampled noise mean to zero within each pair and removes first-order sampling imbalance around the current mean, but it does not eliminate higher-order Monte Carlo variance and is not claimed to be variance-optimal. $K=4$ is fixed for every \Ha{} training run in both tasks, so estimator noise is not a between-condition tuning variable. The numerical self-check in the released \texttt{step2\_train\_muscle\_tanh\_Ha.py} (\texttt{--check --skip\_env}) sweeps $K$ from 1 to 32 and confirms that the estimator mean does not drift with $K$ while its standard deviation shrinks at the expected Monte Carlo rate. The implementation's numerical check compares estimator bias and variance across several $K$ values and verifies the stable Jacobian form; no paper claim relies on treating the four samples as statistical replicates.

\subsection{Release statement and canonical evaluators}
An archival code package accompanies this paper, containing both training implementations, the exact canonical evaluators, analysis scripts, and frozen JSON outputs. A machine-readable manifest maps every reported table to its evaluator revision, checkpoint SHA-256 hash, protocol name, and random-seed convention. The same tagged release will be maintained in a permanent public repository. The canonical evaluators, not training-time rolling diagnostics, are the source of record for reported returns and policy-geometry tables. Where normalization is used, the evaluator restores and freezes the checkpoint-specific observation statistics. The release will also include the common-state and margin scripts so that the state-distribution and threshold robustness claims can be reproduced directly. Large checkpoints and third-party motion-capture assets will be released when licensing permits; otherwise the manifest will provide acquisition instructions and cryptographic hashes.

\section{Reproducibility Notes}
The final analysis uses step-matched checkpoints and versioned canonical evaluators. Several defects discovered during the audit changed earlier reported values or statistical interpretation. Reward-selected checkpoints were not compute matched, and a motor boundary test used the muscle action interval. Beta parameters could fall through a silent fallback, and a mapping could be overwritten by directory order. The first multi-buffer gradient analysis unintentionally replayed the same rollout, because loading the PPO checkpoint reseeded the environments. Finally, Dog evaluator revisions initially mishandled terminal versus periodic checkpoint names. The final scripts add explicit checkpoint metadata, numeric step parsing, regression tests, and buffer fingerprints. These corrections are part of the evidence chain rather than additional experimental conditions.

All experiments were run on Windows 10 with Python 3.9.25, PyTorch 2.8.0 (CUDA 12.8, cuDNN 9.10.2), NumPy 2.0.2, Gymnasium 0.29.1, Stable-Baselines3 2.7.1, MuJoCo 3.3.0, MyoSuite 2.11.6, Shimmy 1.3.0, and SciPy 1.13.1, on a single NVIDIA GeForce RTX 5090 (32\,GiB) with a 32-thread CPU. A machine-generated release manifest reproduces this list together with a full \texttt{pip freeze}, the driver report, per-checkpoint SHA-256 digests, and the resolved Stable-Baselines3 field values read back from the checkpoints rather than assumed. The manifest is produced by a single script and is included with the code archive.

The MyoLeg control loop was verified from the live environment rather than assumed: the physics timestep is 1\,ms and the environment applies a frame skip of 4, giving a 4\,ms control period and a control frequency of 250\,Hz. The Dog-Stand control period is fixed by \texttt{dm\_control} at 0.015\,s (66.67\,Hz).

Training provenance. The no-entropy and \Ha{} seed-0 runs resume 30M-step pilot checkpoints of the same configurations; the symmetric-$\tau$ seeds 1 and 2 continue 100M-step runs to the matched budget; and the filter and override seed-0 policies were trained by earlier single-run drivers that are included in the release alongside the seeded multi-run scripts.

\end{document}